\documentclass[11pt,a4paper]{article}
\usepackage{CJKutf8}
\usepackage{multirow}
\usepackage[dvips]{graphicx}
\usepackage{booktabs}
\usepackage{color}
\usepackage[hyperref]{acl2020}
\usepackage{times}
\usepackage{latexsym}

\usepackage{amsmath}

\usepackage{microtype}

\aclfinalcopy 

\definecolor{x}{RGB}{145,76,152}
\definecolor{y}{RGB}{46,67,146}
\definecolor{n}{RGB}{67,117,92}

\newcommand{\tabincell}[2]{\begin{tabular}{@{}#1@{}}#2\end{tabular}}

\title{Learning to Encode Evolutionary Knowledge for Automatic Commenting Long Novels}

\author{Canxiang Yan\textsuperscript{1}, Jianhao Yan\textsuperscript{1}, Yangyin Xu\textsuperscript{2}, Cheng Niu\textsuperscript{1}, Jie Zhou\textsuperscript{1} \\
        \textsuperscript{1}Pattern Recognition Center, WeChat AI, Tencent Inc, China \\
        \textsuperscript{2}WeChat Reading Center, Tencent Inc, China \\
        \texttt{\{chriscxyan, elliottyan, ayangxu, niucheng, withtomzhou\}@tencent.com} \\}

\date{}

\begin{document}
\maketitle
\begin{abstract}
Static knowledge graph has been incorporated extensively into sequence-to-sequence framework for text generation. While effectively representing structured context, static knowledge graph failed to represent knowledge evolution, which is required in modeling dynamic events. In this paper, an automatic commenting task is proposed for long novels, which involves understanding context of more than tens of thousands of words. To model the dynamic storyline, especially the transitions of the characters and their relations, \textit{Evolutionary Knowledge Graph} (EKG) is proposed and learned within a multi-task framework. Given a specific passage to comment, sequential modeling is used to incorporate historical and future embedding for context representation. Further, a graph-to-sequence model is designed to utilize the EKG for comment generation.
Extensive experimental results show that our EKG-based method is superior to several strong baselines on both automatic and human evaluations.

\end{abstract}

\section{Introduction}
In the past few years, the field of text generation witnesses many significant advances, including but not limited to neural machine translation \cite{Transformer:17,gehring2017convolutional}, dialogue systems \cite{liu-etal-2018-knowledge,zhang2019memory} and text generation \cite{clark-etal-2018-neural,guo2018long}.
By utilizing the power of the sequence-to-sequence (S2S) framework \cite{Sutskever:14}, generation models can predict the next token based on the previous generated outputs and contexts.
However, S2S models are not perfect.
One of the obvious drawbacks is that S2S models tend to be short-sighted on long context and are unaware of global knowledge.
Therefore, how to incorporate global or local knowledge into S2S models has been a long-standing research problem.
\begin{figure}
  \centering
  \includegraphics[width=0.5\textwidth]{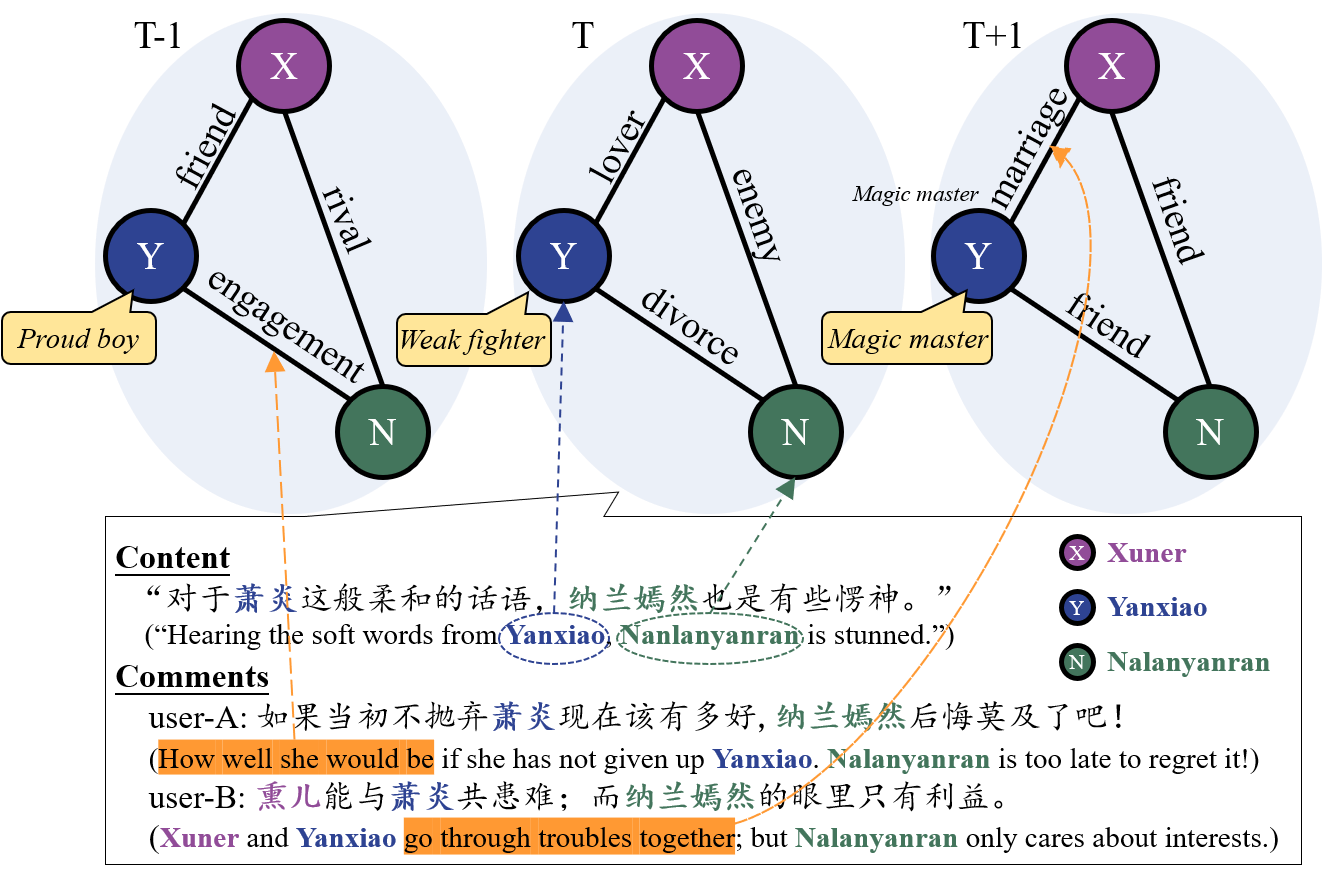}
  \caption{\label{example-table} An example from a novel called ``Fights Break Sphere''. The relations between \textcolor{y}{Yanxiao}, \textcolor{x}{Xuner} and \textcolor{n}{Nalanyanran} are evolutionary. And the characteristic of \textcolor{y}{Yanxiao} changes over time.}
\end{figure}

There lie two different directions to include knowledge into S2S models.
On the one hand, many efforts \cite{zhang-etal-2018-improving,guan2019story,li-etal-2019-incremental} have been taken to address the short-sighted problem in text generation by explicitly modeling unstructured context. Nevertheless, these approaches rely heavily on the quality and scale of unstructured context, and become intractable when applying to scenarios where the context length increases drastically (e.g., commenting a full-length novel).
On the other hand, researchers~\citep{beck-etal-2018-graph,marcheggiani-perez-beltrachini-2018-deep,li-etal-2019-coherent} try to combine knowledge with S2S by employing pre-processed structured data~(e.g., knowledge graph) which naturally avoid the difficulty for context length. However, those models are oriented to static knowledge, and hence hardly model events where temporal knowledge evolution occurs.

The dynamic knowledge evolution is very common in full-length novels. In a novel, a knowledge graph can be constructed by using entities (characters, organizations, locations etc.) as vertices together with the entity relations as edges. Obviously, a single static knowledge graph is hardly to represent the dynamic story line full of dramatic changes. For example, a naughty boy can grow up into a hero, friends may become lovers, etc. In this paper, we proposes \textbf{E}volutionary \textbf{K}nowledge \textbf{G}raph~(\textbf{EKG}) which contains a series of sub-graph for each time step. Figure 1 illustrates EKG for the novel ``Fights Break Sphere''. At three different scenes, ``Yanxiao'', the leading role of the novel, is characterized as ``proud boy'', ``weak fighter'', and ``magic master'' separately. At the same time, the relation between ``Yanxiao'' and ``Xuner'' is evolved from friend into lovers, and finally get married, and the relation between ``Yanxiao'' and ``Nalanyanran'' is changed over time as ``engagement$\rightarrow$divorce$\rightarrow$friend''.

EKG is important for commenting passage of novels since it is the dramatic evolution and comparison in the storyline but not the static facts resonate the readers most. As illustrated in Figure~\ref{example-table}. When commenting passage sampled from the $T$-th chapter of a novel, the user-A refers to a historical fact that "Nalanyanran" has abandoned ``Yanxiao'', while the user-B refers to the future relation between ``Yanxian'' and a related entity ``Xuner'' that they will go through difficulties together. In this paper, EKG is trained within a multi-task framework to represent the latent dynamic context, and then a novel graph-to-sequence model is designed to select the relevant context from EKG for comment generation.

\subsection{Related Work}
graph-to-sequence model has been proposed for text generation.~\citet{song-etal-2018-graph}, ~\citet{beck-etal-2018-graph}, and~\citet{guo-etal-2019-densely} used the graph neural networks to solve the AMR-to-text problem.~\citet{bastings-etal-2017-graph} and ~\citet{GraphSeq2Seq:18} utilized graph convolutional networks to incorporate syntactic structure into neural attention-based encoder-decoder models for machine translation. In comment generation, Graph2Seq~\citep{li-etal-2019-coherent} is proposed to generate comments by modeling the input news as a topic graph. These methods are using static graph, and did not involve the dynamic knowledge evolution.

Recently, more research attention has been focused on dynamic knowledge modeling.~\citet{taheri2019www} used gated graph neural networks to learn the temporal dynamics of an evolving graph for dynamic graph classification.~\citet{KnowEvolve:17, trivedi2018dyrep, kumar2019kdd} learned evolving entity representations over time for dynamic link prediction. Unlike the EKG in this paper, they did not model the embeddings of the relations between dynamic entities.
~\citet{iyyer-etal-2016-feuding} proposed an unsupervised deep learning algorithm to model the dynamic relationship between characters in a novel without considering the entity embedding. Unlike these methods, our EKG-based model represents the temporal evolution of entities and relations simultaneously by learning their temporal embeddings, and hence has an advantage in supporting text generation tasks.

To our knowledge, few studies make use of evolutionary knowledge graph for text generation. This may due to the lack of datasets involving dynamic temporal evolution. We observed that novel commenting need to understand long context full of dramatic changes, and hence build such a dataset by collecting full-length novels and real user comments. The dataset with its EKG will be made publicly available, and more details can be found in Section~\ref{sect:Dataset}.

The main contributions of our work are three-fold:

\begin{itemize}
  \item We build a new dataset to facilitate the research of evolutionary knowledge based text generation.
  \item We propose a multi-task framework for the learning of evolutionary knowledge graph to model the long and dynamic context.
  \item We propose a novel graph-to-sequence model to incorporate evolutionary knowledge graph for text generation.

\end{itemize}

\section{Dataset Development and Evolutionary Knowledge Graph Building}\label{sect:Dataset}
To facilitate the research of modeling knowledge evolution for text generation, we build a dataset called \textit{GraphNovel} by collecting full-length novels and real user comments. Together with the corresponding EKG embeddings, we will make the dataset public available soon.
We detail the collection of the dataset below.

\subsection{Data collection}
\label{sub:collect}
The data is collected from well-known Chinese novel websites. To increase the diversity of data, top-1000 hottest novels are crawled with different types including science fiction, fantasy, action, romance, historical, and so on. Then we filter out novels due to the following 3 considerations: 1) the number of chapters is less than 10, 2) few entities are mentioned and 3) lack of user comments. Each remained novel includes chapters in chronological order, a set of user-underlined passages, and user comments for the passages.

Then, we use the lexical analysis tool~\citep{jiao2018LAC} to recognize three types of entities (persons, organizations, locations) from each novel. Due to many of the nickname in novels, the identified entities from the tool contains much noise. To improve the knowledge quality, human annotators are asked to verify the entities, and add missing ones. Then all the paragraphs containing mentions of two entities are identified, and will later serve as a representation of the entity relations at that specific time step.

As for the highlighted novel passage and user comments, three criteria are used to select high quality and informative data: 1) The selected passage must contain at least one entity; 2) the selected passage must be commented by at least three users; 3) comments related to the same passage are ranked according to the upvotes, and the bottom 20\% are dropped. Notably, those highlighted passages have a degree of redundancy because users tend to highlight and comment passages at very similar positions. Thus we merge passages which have more than 50\% overlapping rate. This operation can effectively reduce the quantity of passages by 30\%.


\subsection{Core Statistics}
The dataset contains 203 novels, 349,695 highlighted passages and 3,136,210 comments totally. Due to diverse genre of novels in our dataset, the number of entities and relations per novel varies widely with range [$10^2$, $10^5$]. And the number of comments per passage changes a lot with a range of [$3$, $2\times{10^3}$], because it depends on how interesting the corresponding passage is.

\begin{figure*}[t]
  \centering
  \includegraphics[width=\textwidth]{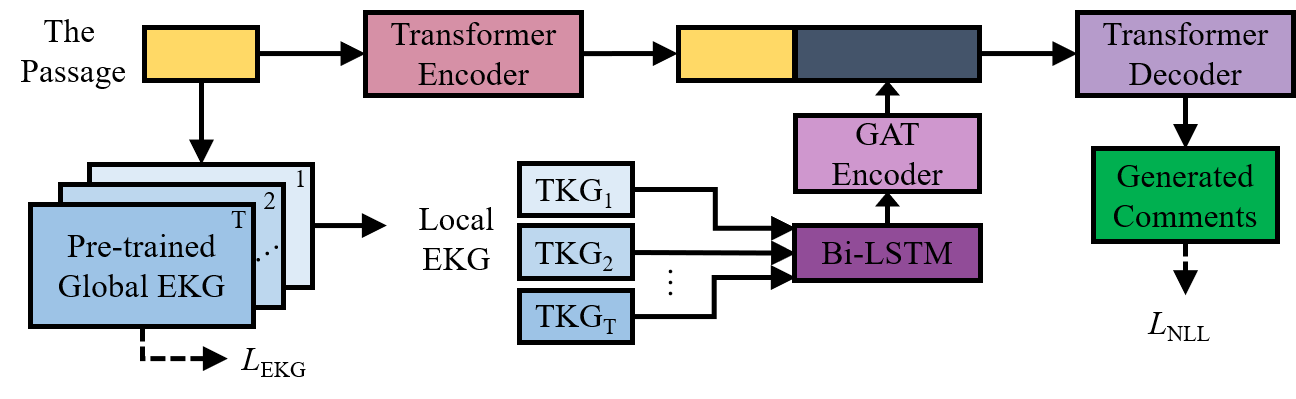}
  \caption{Architecture of our model. First, the EKG is trained under a multi-task learning framework. Then a graph-to-sequence model is traind to utilize the EKG for text generation.}\label{fig:model}
\end{figure*}

We partitioned the dataset into non-overlapping training, validation, and test portions, along novels (See Table~\ref{dataset-table} for detailed statistics). Five most relevant comments for each passage in validation and test set are selected by human annotators. While the comments in train set are all preserved in order to ensure its flexible use.


\begin{table}[t]
\centering
\begin{tabular}{l|r|r|r}
\hline
 & \textbf{train} & \textbf{valid} & \textbf{test} \\
\hline
{\small{\# novels}} & 173 &  10 & 20 \\
\hline
{\small{\# passages}} & 324,803 & 7,976 & 16,916 \\
\hline
{\small{\# comments}} & 3,011,750 & 39,880 & 84,580 \\
\hline
\tabincell{l}{\small{Avg. length}\\\small{~~of context}} & 520,571 & 305,492 & 277,847 \\
\hline
{\small{Avg. \# entities}} & 383.4 & 720.6 & 281.7 \\
\hline
{\small{Avg. \# relations}} & 9,013 & 1,9919 & 7,084 \\
\hline
\tabincell{l}{\small{Avg. \# comments}\\\small{~~per passage}} & 9.3 & 5.0 & 5.0 \\
\hline
\tabincell{l}{\small{Avg. \# entities}\\\small{~~per passage}} & 2.6 & 3.1 & 3.4 \\
\hline
\tabincell{l}{\small{Avg. \# relations}\\\small{~~per passage}} & 4.4 & 9.2 & 11.0 \\
\hline
\end{tabular}
\caption{\label{dataset-table} Statistics of Dataset. }
\end{table}

\subsection{Build up Evolutionary Knowledge Graph}
\label{sub:collect}

Then for each novel, we build up a knowledge graph which consists of a sequence of sub-graphs. Obviously, it is sensible to build up a sub-graph for each import scene of the novel, and build up more sub-graphs around the critical transitions in the stroryline. In this paper, we assume each chapter usually represents an integral scene, and hence build a sub-graph for each chapter of the novel.

Then for each chapter, the entities being mentioned in the chapter are the vertices of the corresponding sub-graph. And if a paragraph consists two of the entities, an edge is created between the two entities. In such a way, a sequence of sub-graphs are constructed, and form our EKG. In the next section, we will formulate the embedding computation of the EKG, and its application for comment generation.

\section{Model Formalism}\label{sec:formalism}
In this section, we formulate our approach in details. First, the training of EKG embedding is presented. Then a graph-to-sequence model is shown to utilize the EKG for comment generation. The architecture of the model is shown in Figure~\ref{fig:model}.

\subsection{Definition}
For a novel with $n_e$ entities and $n_r$ relations, define a global evolutionary knowledge graph: $G_{ekg}^{global} = \{G(t)\}|_{t:1\rightarrow{T}}$,
where $T$ is the number of chapters\footnote{We will cluster successive chapters into a longer one if they are too short.} (or time periods); $G(t)=\langle{V(t), E(t)}\rangle$, is a temporal knowledge graph of the chapter $t$; $V(t)=\{v_1(t), v_2(t),...,v_{n_e}(t)\}$ is the set of vertices and $E(t)=\{e_1(t), e_2(t),...,e_{n_r}(t)\}$ is the set of edges between two vertices.

Given a passage $C$ from the chapter $t$ with $c_e$ entities and $c_r$ relations, a local EKG related to it can is a sub-graph of the global EKG: $G_{ekg}^{local}=\{G_c(t)\}|_{t:1\rightarrow{T}}$, where $G_c(t)=\langle{V_c(t), E_c(t)}\rangle$; $V_c(t)$ is a subset of $V(t)$ with size of $c_e$ and $E_c(t)$ is a subset of $E(t)$ with size of $c_r$. Then the local EKG of the passage is a sequence of local temporal knowledge sub-graphs with $T \times c_e$ vertex embeddings and $T \times c_r$ edge embeddings.
\subsection{EKG Embedding Training}
Inspired by the consistent state-of-the-art performance in language understanding tasks, we use off-the-shelf Chinese BERT model~\citep{devlin-etal-2019-bert} to calculate the initial semantic representation of sentences. Considering the fact that entities are either out-of-vocabulary or associated with special semantics within the novel context, we propose the following algorithm to jointly learn entity and relation embeddings in EKG:

\paragraph{Vertex embedding learning.} The passages containing mentions of entity $v$ will contribute to learn the embedding of $v$. Specifically, the $i$-th passage is tokenized while the entity mention $v_i$ is masked with token ``[MASK]''. Then resulted tokens are fed into the pre-trained Chinese BERT model, and $f_{v_i}$ is obtained as the output feature corresponding to the mask token.
Within the chapter $t$, there exists $N_{v}^{t}$ sentences containing vertices $v$. The embedding of $v$ is learned by optimizing the following softmax loss summation which models the probabilities to predict the masked entities as $v$.
\begin{equation}\label{node_learn}
  L_{v}^{t} = -\sum_{i=1}^{N_{v}^{t}}{\log{p_{v_i}^{t}}}
\end{equation}
\begin{equation}\label{node_softmax}
  p_{v_i}^{t} = softmax(W_{v}^{t}\cdot{f_{v_i}})
\end{equation}
where $W_v^t$ is learnable parameter and denotes the embedding of $v$ from the chapter $t$. Usually the semantic representations of entities change smoothly over time, so we propose a temporally smoothed softmax loss to retain the similarity of entity embeddings from successive time periods:
\begin{equation}\label{node_improve}
   \tilde{L}_{v}^{t} = -\sum_{i=1}^{N_{v}^{t}}{(\lambda{_0}\log{p_{v_i}^{t-1}}+\lambda{_1}\log{p_{v_i}^{t}}+\lambda{_2}\log{p_{v_i}^{t+1}})}
\end{equation}
where $\lambda_0$, $\lambda_1$ and $\lambda_2$ are smooth factors; and only valid probabilities are included when $t=1$ or $T$. Then the overall loss for all time periods and all vertices is:
\begin{equation}\label{node_overall}
  L_{vertex} = \sum_{t=1}^{T}\sum_{v=1}^{n_e}{\tilde{L}_{v}^{t}}
\end{equation}

\begin{figure}[t]
  \centering
  \includegraphics[width=0.5\textwidth]{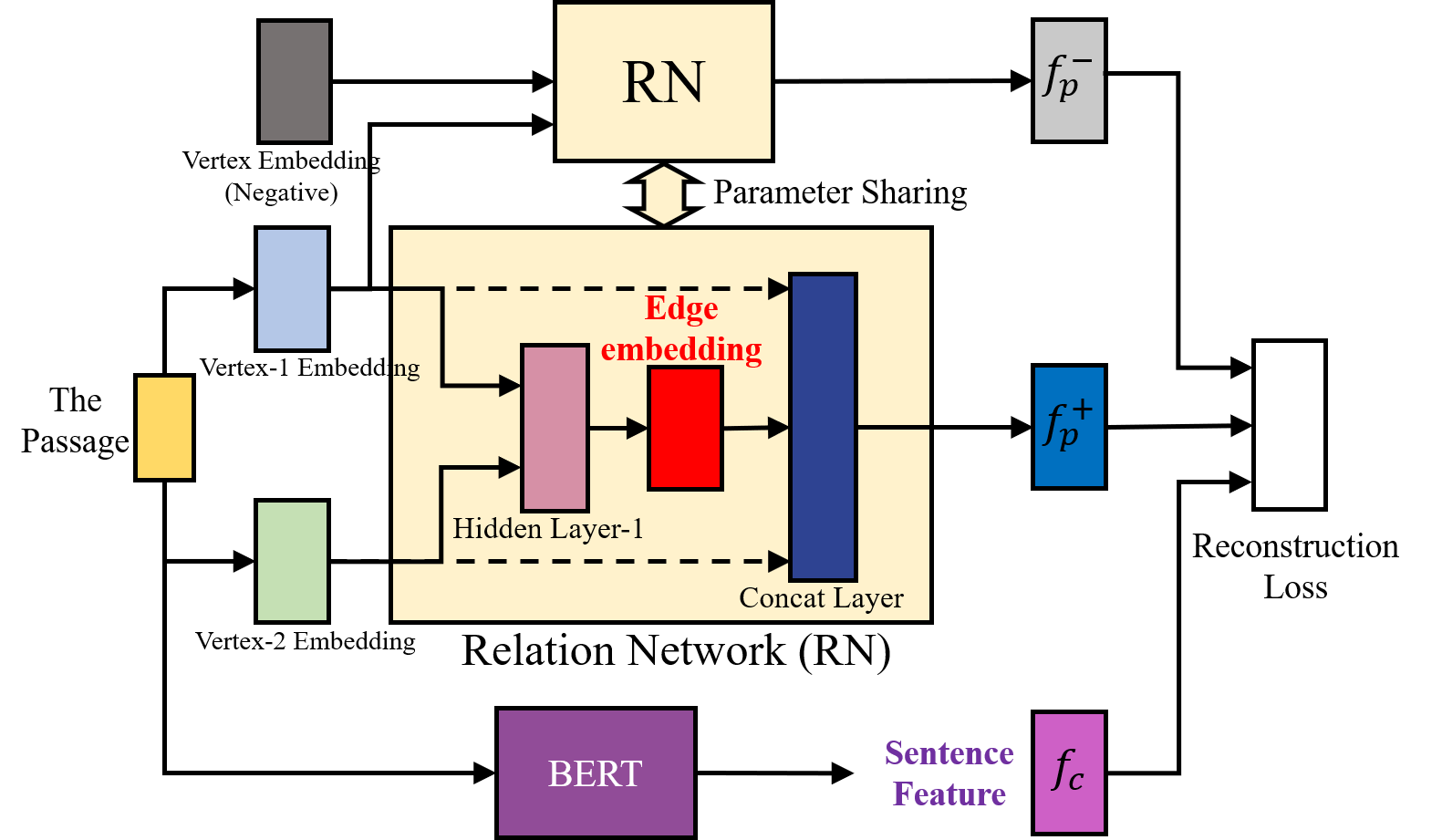}
  \caption{Relation Network with a reconstruction loss.The edge embedding is shown in \textcolor{red}{red color}.}\label{fig:rn}
\end{figure}

\paragraph{Edge embedding learning.} Since the number of relations equal the number of co-occurrence for any two entities, it is infeasible to employ an embedding matrix to model the relation. Therefore, a \textit{Relation Network} (RN) is proposed to learn the edge embeddings in the TKGs as shown in Figure~\ref{fig:rn}. Specifically, the RN takes two vertex embeddings as input, and feed them into the first hidden layer to obtain the embedding of the edge $r$. Then the embeddings of two vertices and one edge are concatenated and fed into second hidden layer to reconstruct the sentence. We also use the pre-trained BERT~\citep{devlin-etal-2019-bert} to obtain the representation $f_c$ for the whole sentence. The $f_c$ is taken from the final hidden state corresponding to ``[CLS]'' token because it aggregates sequence representation.

A reconstruction loss applied to the network is optimized to jointly learn RN and edge embeddings:
\begin{equation}\label{edge_l0}
  L_r = \max(d(f_{p^+}, f_c)- d(f_{p^-}, f_c) + \alpha, 0)
\end{equation}
where $p$ stands for a pair of vertices; $p^+$ represents a positive pair with two vertex related to the edge, and $p^-$ represents a negative pair with one vertex related to the edge and the other unrelated. The overall loss for learning edge embedding is:
\begin{equation}\label{edge_loss}
  L_{edge} = \sum_{r=1}^{N_r}L_r
\end{equation}

Combining the vertex and edge loss above, our final multi-task loss is:
\begin{equation}\label{tkg_loss}
  L_{EKG} = L_{vertex} + \lambda_{r}L_{edge}
\end{equation}
where $\lambda_{r}$ is a hyperparameter to be tuned.

\subsection{Graph-to-sequence modeling}
After EKG embedding learning, we propose a graph encoder to utilize the embeddings of the EKG for comment generation. From the learned $G_{ekg}^{local}$, we can obtain vector sequences $V_i=\{v_i(1),..,,v_i(2), v_i(T)\}$ for each vertex, and $E_i=\{e_i(1),e_i(2),...e_i(T)\}$ for each edge. All these sequences are fed into Bi-LSTM to integrate information from all time periods. Then final representation of the vertices and edges are taken from the final hidden state of Bi-LSTM corresponding to the time step $t$.

Further, our graph encoder employs graph convolutional networks~\citep{GNN:18} to aggregate the structured knowledge from the EKG and then is combined into a widely used encoder-decoder framework~\citep{Transformer:17} for generation.

our graph encoder is based on the implementation of GAT~\citep{v2018graph}. The input to a single GAT layer is a set of vertices and edge features, denoted as $\mathbf{F}_v=\{\vec{v}_1, \vec{v}_2,..., \vec{v}_{c_e}\}$, and $\mathbf{F}_r=\{\vec{r}_1, \vec{r}_2,..., \vec{r}_{c_r}\}$, where ${c_e}$ is the number of vertices, and ${c_r}$ is the number of edges from the passage. The layer produces a new set of vertex features, $\mathbf{F'}_v=\{\vec{v'}_1, \vec{v'}_2,..., \vec{v'}_{c_e}\}$ as its output.

In order to aggregate the structured knowledge from input features and transform them into higher-level features, we then perform self-attention on both the vertices and the edges to compute attention coefficients
\begin{equation}\label{gat_atten1}
  \alpha_{ij}^e = g(\mathbf{W}\vec{v}_i, \mathbf{W}\vec{v}_j)
\end{equation}

\begin{equation}\label{gat_atten2}
  \alpha_{ij}^r = h(\mathbf{W}\vec{v}_i, \mathbf{W}\vec{r}_{ij})
\end{equation}
where $\mathbf{W}$ is learnable parameter; $g$ and $h$ are mapping functions; $\alpha_{ij}^e$ and $\alpha_{ij}^r$ indicate the importance of neighbor features to vertex $i$; they are normalized using softmax.

Once obtained, the normalized attention coefficients are used to compute a linear combination of the features corresponding to them, to serve as the final output features for every vertex:
\begin{equation}\label{gat_output}
  \vec{v'}_i=\sum_{j\in{\mathcal{N}_i}}\alpha_{ij}^e\mathbf{W}\vec{v}_j + \alpha_{ij}^r\mathbf{W}\vec{r}_{ij}
\end{equation}

Then the graph encoder is combined into the encoder-decoder framework~\citep{Transformer:17}, in which a self-attention based encoder is used to encode the passage. To aggregate structured knowledge, the encoding of all vertices from graph encoder are concatenated with the output of the passage encoder, and fed into a Transformer decoder for text generation. The whole graph-to-sequence model is trained to minimize the negative log-likelihood of the related comment.
\section{Experiment}
In this section, we first introduce the experimental details, and then present results from automatic and human evaluations.
\subsection{Details}
We train all the models using training set and tune hyperparamters based on validation set. The automatic and human evaluations are carried out based on test set.
During the training of the EKG, we first learn the vertex embeddings and fix them during the subsequent training of edge embeddings. Our GAT-based graph encoder is based on the entities from the passage. we keep $K$ entities for each passage: if the number of entities\footnote{Note that the number will not be zero because the passage contains as least one entity in our dataset.} is smaller than $K$, we use breadth-first searching on the global graph to fill the gap, otherwise we filter the low-order entities according to entity frequency. We set $K$ to 5 which is selected by validation. Label smoothing is used in the smoothed softmax loss. We denote our full model as ``\textbf{EKG+GAT(V+E)}'', its variant which only use the first term in Equ.~\ref{gat_output} as ``\textbf{EKG+GAT(V)}'' and name the other variant ``\textbf{EKG}'', which does not use GAT-based graph encoder and feeds the encoding of vertices into the Transformer decoder directly.
\subsection{Hyperparameters}
In our model, we set $\lambda_1=0.5$, $\lambda_2=1.0$, $\lambda_3=0.3$ for the smoothed softmax loss; set $\alpha$ to 0.0 for the reconstruction loss and $\lambda_r$ to 1.0 for the multi-task loss; the number of self-attention layers in our passage encoder is 6; the number of Bi-LSTMs is 2 and the length of its hidden state is 768; and the GAT-based graph encoder has two layers. To stabilize the training, we use Adam optimizer~\citep{adam:14} and follow the learning rate strategy in~\citet{klein-etal-2017-opennmt} by increasing the learning rate linearly during the first $5000$ steps for warming-up and then decaying it exponentially. For inference, the maximum length of decoding is 50; beam searching is used with beam size 4 for all the models.
\subsection{Evaluation metrics}
we use both automatic metrics and human evaluations to evaluate the quality of generated novel comments.

\noindent{\textbf{Automatic metrics}}:~1) \textbf{BLEU} is commonly employed in evaluating translation systems. It is also introduced into comment generation task~\citep{qin-etal-2018-automatic, yang-etal-2019-read}. we use $multi{-}bleu.perl$\footnote{https://github.com/moses-smt/mosesdecoder/blob/master/scripts/generic/multi-bleu.perl} to calculate the BLEU score. 2) \textbf{ROUGE-L}(\citep{lin-2004-rouge}) uses longest common subsequence to calculate the similar score between candidates and references. For calculation, we use a python package called $pyrouge$\footnote{https://pypi.org/project/pyrouge/}. These metrics also support the multi-reference evaluations on our dataset.

\noindent{\textbf{Human evaluations}}:~1) \textbf{Relevance}: This metric evaluates how relevant is the comment to the passage. It measures the degree that the comment is about the main storyline of the novel.2) \textbf{Fluency}: This metric evaluates whether the sentence is fluent and judges whether the sentence follows the grammar and whether the sentence has clear logic. 3) \textbf{Informativeness}: This metric evaluates how much structured knowledge the comment contains. It measures whether the comment reflects the evolution of entities and relations, or is just a general description that can be used for many passages. All these metrics have three gears, the final scores are projected to 0$\sim$3.

\subsection{Baseline Models}
We describe three kinds of models used as baselines. All the baselines are implemented according to the related works and tuned on the validation set.
\begin{table*}[!t]
\centering
\begin{tabular}{lccccc}
\hline \textbf{Model} & \textbf{BLEU} &\textbf{ROUGE-L}
& \textbf{Relevance} & \textbf{Fluency} & \textbf{Informativeness}\\ \hline
\textbf{Seq2Seq}\citep{qin-etal-2018-automatic} & 2.59 & 14.71 & 0.12 & 1.71 & 0.09\\
\textbf{Attn}\citep{qin-etal-2018-automatic} & 3.71 & 16.44 & 0.34 & 1.70 & 0.33 \\
\textbf{Trans}\citep{Transformer:17} & 6.11 & 19.21 & 0.57 & 1.62 & 0.58 \\
\textbf{Trans.+CTX}\citep{zhang-etal-2018-improving} & 6.52 & 19.11 & 0.68 & 1.68 & 0.67\\
\textbf{Graph2Seq}\citep{li-etal-2019-coherent} & 4.93 & 16.91 & 0.35 & 1.69 & 0.31 \\
\textbf{Graph2Seq++}\citep{li-etal-2019-coherent} & 5.56 & 17.51 & 0.85 & 1.67 & 0.60\\
\hline
\textbf{EKG} & 6.59 & 20.00 & 0.81 & \textbf{1.83} & 0.64 \\
\textbf{EKG+GAT(V)} & 6.72 & 20.09 & 0.88 & 1.77 & 0.70 \\
\textbf{EKG+GAT(V+E)} & \textbf{7.01} & \textbf{20.10} & \textbf{0.89}  & 1.74 & \textbf{0.75}  \\
\hline
\textbf{Human Performance} & 100 & 100 & 1.09 & 1.85 & 1.04\\
\hline

\end{tabular}
\caption{\label{metric-table} Automatic metrics and human evaluations. }
\end{table*}
\begin{itemize}
  \item \textbf{Seq2Seq models}~\citep{qin-etal-2018-automatic}:~those models generate comments for news either from the title or the entire article. Considering there are no titles in our dataset, We compare two kinds of models from their work. 1) \textbf{Seq2Seq:} it is a basic sequence-to-sequence model~\citep{Sutskever:14} that generate comments from the passage; 2) \textbf{Attn:} sequence-to-sequence model with an attention mechanism\citep{Bahdanau:14}. For the input of the attention model, we append the related entities to the back of the passage.

  \item \textbf{Self-attention models}:~our model includes a graph encoder to encode knowledge from graph, and a passage encoder use multiple self-attention layers. To show the power of graph encoder, we use the encoder-decoder framework (\textbf{Trans.})~\citep{Transformer:17} for passage-based comparison. Also we introduce an improved Transformer~\citep{zhang-etal-2018-improving} with a context encoder to represent document-level context and denote it \textbf{Trans.+CTX}. For the context input, we use up to 512 tokens before the passage as context.

  \item \textbf{Graph2Seq}~\citep{li-etal-2019-coherent}:~this is a graph-to-sequence model that builds a static topic graph for the input and generates comments based on representations of entities only. A two-layer transformer encoder is used in their work. For fair comparison, we use 6-layer transformer encoder to replace the original and denote the new model as \textbf{Graph2Seq++}.
\end{itemize}

\subsection{Evaluation results}
Table~\ref{metric-table} shows the results of both automatic metrics and human evaluations.

In automatic metrics, our proposed model has best BLEU and ROUGE-L scores.
For BLEU, our full model EKG+GAT(V+E) achieves 7.01 score, which is 0.59 higher than that of the best baseline Trans.+CTX.
The Graph2Seq++ has a BLEU score 5.56 which is obviously lower than the EKG+GAT(V+E). The main reason is that the Graph2Seq++ is based on static graph and cannot make use of the dynamic knowledge.
For Rouge-L, our models all have ROUGE-L scores higher than 20\%, and is 0.79\% slightly better than that of Trans., which is the best among all baselines; the ROUGE-L score of Trans. is higher than of the Trans.+CTX, which is opposite of that in BLEU.

In human evaluations, we randomly select 100 passages from the test set and run all the models in Table~\ref{metric-table} to generate respective comments. We also provide one user comment for each passage to get evaluations of human performance. All these passage-comment pairs are labeled by human annotators.
In relevance metric, our full model EKG+GAT(V+E) has better relevance score than all the baselines. It means our model can generate more relevant comments and better reflect the main storyline of the novel. For all this, there still exists significant gaps when compared to the human performance.
In fluency and informativeness metrics, our EKG+GAT(V+E) model has achieves higher score compared to all baselines. It illustrates that the generated comments by our proposed model are more fluent and contains more attractive information.

\begin{table*}[!t]\small
\centering
\begin{tabular}{|l|}
\hline
\begin{CJK*}{UTF8}{gkai}\tabincell{l}{\textbf{P1:}~\textbf{这人家姓曾，住在县城以南一百三十里外的荷叶塘都。}\\
(This family, surnamed Zeng, lives in Heyetangdu, 130 miles \textcolor{orange}{south of the county}.)
}\end{CJK*} \\
\hline
\begin{CJK*}{UTF8}{gkai}\tabincell{l}{\textbf{T1:}~原来是这么来的。~(That's how it turned out.)}\end{CJK*} \\
\begin{CJK*}{UTF8}{gkai}\tabincell{l}{\textbf{G1:}~这个地方呀!~(This is the place !)}\end{CJK*} \\
\begin{CJK*}{UTF8}{gkai}\tabincell{l}{\textbf{E1:}~他一直思念着家里人。~(He has been \textcolor{cyan}{missing his family}.)}\end{CJK*}~\textcolor{blue}{\textbf{[P2]}}\\
\hline
\begin{CJK*}{UTF8}{gkai}\tabincell{l}{\textbf{P2:}~\textbf{国藩今日乃戴孝之身，老母并未安葬妥帖，怎忍离家出山?}\\
(Today Guofan is wearing mourning. \textcolor{cyan}{My mother hasn't been buried yet}. How can I leave home ?)
}\end{CJK*} \\
\hline
\begin{CJK*}{UTF8}{gkai}\tabincell{l}{\textbf{T2:}~真是一个聪明人~(He is so clever.)\\}\end{CJK*} \\
\begin{CJK*}{UTF8}{gkai}\tabincell{l}{\textbf{G2:}~老太太也是个好人~(The old lady is also a good person.)\\}\end{CJK*} \\
\begin{CJK*}{UTF8}{gkai}\tabincell{l}{\textbf{E2:}~这个时候的国家已经有了变化~(\textcolor{green}{The country is changing} at this time.)}\end{CJK*} ~\textcolor{blue}{\textbf{[P3]}}\\
\hline
\begin{CJK*}{UTF8}{gkai}\tabincell{l}{\textbf{P3:}~\textbf{面临大敌，曾暗自下定决心，一旦城破，立即自刎，追随塔齐布、罗泽南于地下。}}\end{CJK*} \\
(\textcolor{green}{Facing the enemy}, Zeng made up his mind to \textcolor{red}{commit suicide as soon as the city broke}, following Ta Qibu and Luo Zenan.)\\
\hline
\begin{CJK*}{UTF8}{gkai}\tabincell{l}{\textbf{T3:}~这就是原来战争的样子}\end{CJK*} (This is what the war looks like.)\\
\begin{CJK*}{UTF8}{gkai}\tabincell{l}{\textbf{G3:}~一个人的命运总是如此}\end{CJK*}(One's destiny is always like this.)\\
\begin{CJK*}{UTF8}{gkai}\tabincell{l}{\textbf{E3:}~自立于南城，自破而立~(He established himself in \textcolor{orange}{the south of the country} throught constant breakthroughs.)}\end{CJK*}~\textcolor{blue}{\textbf{[P1]}}\\
\hline
\begin{CJK*}{UTF8}{gkai}\tabincell{l}{\textbf{P4:}~\textbf{曾国藩的脸上露出一丝浅浅的笑意，头一歪，倒在太师椅上.}\\
(Zeng Guofan smiled slightly. His head tilted and fell on the chair.)
}\end{CJK*} \\
\hline
\begin{CJK*}{UTF8}{gkai}\tabincell{l}{\textbf{T4:}~这一段描写真的很有画面感~(This description is really picturesque.)}\end{CJK*} \\
\begin{CJK*}{UTF8}{gkai}\tabincell{l}{\textbf{G4:}~这个人的心思缜密~(This man is very thoughtful.)}\end{CJK*} \\
\begin{CJK*}{UTF8}{gkai}\tabincell{l}{\textbf{E4:}~他一生忠君为国，就这样走了~(He was \textcolor{red}{loyal to his country} all his life; he is gone.)}\end{CJK*}~\textcolor{blue}{\textbf{[P3]}}\\
\hline
\end{tabular}
\caption{\label{case-table}Comments generated by Trans.+CTX~(\textbf{T}), Graph2Seq++~(\textbf{G}) and our EKG+GAT(V+E)~(\textbf{E}). The passages (i.e., P1, P2, P3, P4) are extracted from the same novel called \emph{Zeng Guofan}. We highlight the passage corresponding to the generated comment from our model~\textbf{E} with \textcolor{blue}{blue color}. Moreover, the relevant fragments are marked with a same color.}
\end{table*}
\begin{figure}
  \centering
  \includegraphics[width=0.5\textwidth]{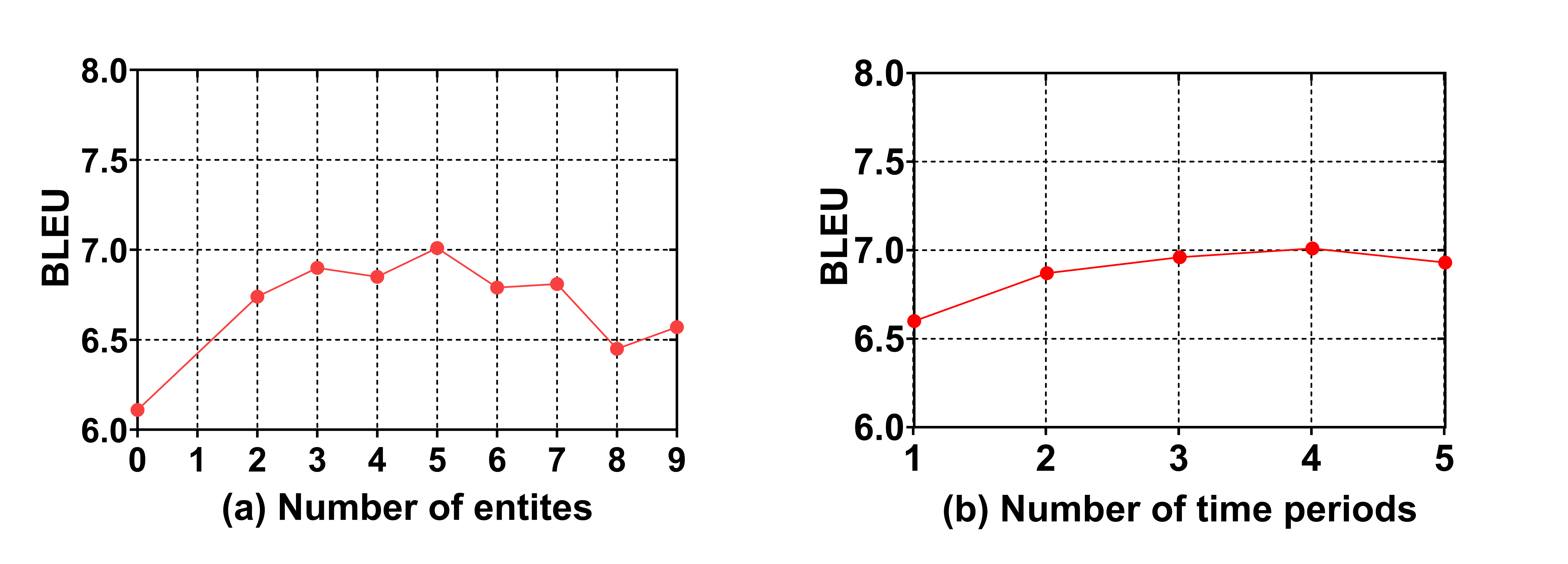}
  \caption{Ablation results about number of entities (a) and number of time periods (b).}\label{fig:ablation}
\end{figure}
\subsection{Analysis and Discussion}
\paragraph{Ablation study:} we compare the results of EKG, EKG+GAT(V), and EKG+GAT(V+E). The EKG, which does not use graph encoder, can achieve 6.59 BLEU score, which is 1.03 higher than that of Graph2Seq++. Then the BLEU score can be further improved to 6.72 by introducing a vertex-only variant EKG+GAT(V). Comparing EKG+GAT(V) and EKG+GAT(V+E) to the EKG, the BLEU scores increase 0.13 and 0.42 respectively; it indicates the usefulness of the graph encoder and that the evolutionary knowledge from edges can be treated as a good supplement to that of vertices. In human evaluations, EKG+GAT(V+E) and EKG+GAT(V) have higher relevance and informativeness scores than that of the EKG. It also indicates that the graph encoder can effectively utilize the evolutionary knowledge from vertices and edges, and make the generated comments more relevant and informative.
\paragraph{Analysis of the number of entities:} the corresponding local EKG is constructed based on the entities from the passage. To explore the influence of the number ($N$) of entities, we report BLEU scores of our full model based on different number\footnote{We do not report the BLEU score of the full model when $N=1$ because there are no edges included.} in Figure~\ref{fig:ablation}(a). The best BLEU score is achieved at $N=5$. The BLEU score at $N=0$ belongs to the Transformer(Trans.). And our full model is robust to the number of entities because the BLEU scores are stable when N is in the range $[2, 7]$.
\paragraph{Analysis of the number of time periods:} we also report the BLEU scores under the different number of time periods in Figure~\ref{fig:ablation}(b). Our full model achieves the best BLEU score of 7.01 at $N=4$, which is 0.41 higher than that of the static graph at $N=1$. It illustrates that the dynamic knowledge is useful for improving the performance.

\paragraph{Case study:} we provide a case study here. Four passages that need to be commented are extracted from a novel chronologically and shown in Table~\ref{case-table}. For comparison, we use \textbf{T}rans.+CTX and \textbf{G}raph2Seq++, which have the best relevance and informativeness scores among baselines respectively. To start with, within each case, we find that the generated comments from our model are more informative, while the generated outputs from other models tend to be general or common replies, which proves the effectiveness of our knowledge usage.

From another perspective, we observe that our model can make use of the dynamics of knowledge. Let us take a look at \textbf{P3}, our generated comment describes that \textit{Zeng Guofan} was born in the south of the country, which is in accordance with the passage described in \textbf{P1}. Similar interactions can be found in all four cases, which support our claims above.
\section{Conclusion}
In this paper, we propose to encode evolutionary knowledge for automatic commenting long novels. We learn an \textit{Evolutionary Knowledge Graph} under a multi-task framework and then design a graph-to-sequence model to utilize the EKG for generating comments. In addition, we collect a new generation dataset called \textit{GraphNovel} to advance the corresponding research. Experimental results show that our EKG-based model is superior to several strong baselines on both automatic metrics and human evaluations. In the future, we plan to develop new graph-based encoders to generate personalized comments with this dataset.

\bibliography{anthology,acl2020,mybib}
\bibliographystyle{acl_natbib}

\end{document}